\documentclass[11pt,a4paper]{article}
\usepackage[hyperref]{ranlp2023}
\usepackage{times}
\usepackage{latexsym}

\usepackage{graphicx}
\usepackage[acronym]{glossaries}

\usepackage{microtype}

\aclfinalcopy 

\title{Bipol: Multi-axes Evaluation of Bias with Explainability in Benchmark Datasets}

\date{}

\author{\\Tosin Adewumi*\textsuperscript{+}, Isabella  Södergren\textsuperscript{++}, Lama Alkhaled\textsuperscript{+}, Sana Sabah Sabry\textsuperscript{+}, \\ Foteini Liwicki\textsuperscript{+} and Marcus Liwicki\textsuperscript{+}
\\
\textsuperscript{+}Machine Learning Group, EISLAB, \textsuperscript{++}Digital Services and Systems
\\ Luleå University of Technology, Sweden \\
\large
{\fontfamily{pcr}\selectfont
\textsuperscript{+}firstname.lastname@ltu.se, \textsuperscript{++}isasde-5@student.ltu.se}
}

\makeglossaries

\newacronym{ood}{OOD}{out-of-distribution}
\newacronym{nlp}{NLP}{Natural Language Processing}
\newacronym{ner}{NER}{Named Entity Recognition}
\newacronym{sa}{SA}{Sentiment Analysis}
\newacronym{bow}{BoW}{bag-of-words}
\newacronym{cbow}{CBoW}{continuous Bag-of-Words}
\newacronym{sltc}{SLTC}{Swedish Language Technology Conference}
\newacronym{ann}{ANN}{artificial neural network}
\newacronym{nn}{NN}{neural network}
\newacronym{lstm}{LSTM}{Long Short Term Memory Network}
\newacronym{sota}{SotA}{state-of-the-art}
\newacronym{nlg}{NLG}{Natural Language Generation}
\newacronym{mwe}{MWE}{Multi-Word Expression}
\newacronym{cnn}{CNN}{Convolutional Neural Network}
\newacronym{sw}{SW}{Simple Wiki}
\newacronym{mt}{MT}{Machine Translation}
\newacronym{gdc}{GDC}{Gothenburg Dialog Corpus}
\newacronym{t5}{T5}{Text-to-Text-Transfer Transformer}
\newacronym{roberta}{RoBERTa}{Robustly optimized BERT approach}
\newacronym{bert}{BERT}{Bidirectional Encoder Representations from Transformers}
\newacronym{mcc}{MCC}{Matthews Correlation Coefficient}
\newacronym{ai}{AI}{artificial intelligence}
\newacronym{xai}{XAI}{explainable artificial intelligence}
\newacronym{lime}{LIME}{Local Interpretable Model-agnostic Explanations}
\newacronym{bilstm}{Bi-LSTM}{Bi- Directional Long Short Term Memory Network}
\newacronym{rnn}{RNN}{Recurrent Neural Network}
\newacronym{ml}{ML}{machine learning}
\newacronym{hasoc}{HASOC}{hate speech and offensive content}
\newacronym{olid}{OLID}{offensive language identification dataset}
\newacronym{ig}{IG}{Integrated Gradient}
\newacronym{shap}{SHAP}{SHapley Additive exPlanations}
\newacronym{hs}{HS}{hate speech}
\newacronym{trac}{TRAC}{Trolling, Aggression and Cyberbullying}
\newacronym{hso}{HSO}{hate speech and offensive}
\newacronym{ooc}{OOC}{out-of-class}
\newacronym{os}{OS}{operating system}
\newacronym{lr}{LR}{learning rate}
\newacronym{multiwoz}{MultiWOZ}{Multi-Domain Wizard-of-Oz}
\newacronym{dialogpt}{DialoGPT}{Dialogue Generative Pre-trained Transformer}
\newacronym{bold}{BOLD}{Bias in Open-Ended Language Generation Dataset}
\newacronym{cuad}{CUAD}{Contract Understanding Atticus Dataset}
\newacronym{ledgar}{LEDGAR}{Labeled Electronic Data Gathering, Analysis, and Retrieval system}
\newacronym{genbit}{GenBiT}{
Gender bias in text toolkit}
\newacronym{imdb}{IMDB}{
Internet Movie Database}
\newacronym{mdgender}{MDGender}{
Multi-Dimensional Gender}
\newacronym{sbic}{SBICv2}{Social Bias Inference Corpus v2}
\newacronym{tp}{tp}{true positives}
\newacronym{tn}{tn}{true negatives}
\newacronym{fp}{fp}{false positives}
\newacronym{fn}{fn}{false negatives}
\newacronym{bat}{BAT}{Bias-Aware Thresholding}
\newacronym{tf}{TF}{term frequency}
\newacronym{mab}{MAB}{multi-axes bias dataset}
\newacronym{squad}{SQuADv2}{Stanford Question Answering Dataset}
\newacronym{qa}{QA}{question-answering}
\newacronym{boolq}{Boolq}{Boolean Question}
\newacronym{cb}{CB}{CommitmentBank}
\newacronym{wsc}{WSC}{Winograd Schema Challenge}
\newacronym{axg}{AXg}{Winogender diagnostic}
\newacronym{rte}{RTE}{Recognising Textual Entailment}
\newacronym{llm}{LLM}{large language models}
\newacronym{mt5}{mT5}{multilingual T5}
\newacronym{nmt}{NMT}{neural machine translation}
\newacronym{pii}{PII}{Personal identifiable information}

\begin{document}

\maketitle
\textbf{Caution: This paper contains examples, from datasets, of what some may consider as stereotypes or offensive text.}

\vspace{10pt}

\begin{abstract}
We investigate five English \acrshort{nlp} benchmark datasets (on the superGLUE leaderboard) and two Swedish datasets for bias, along multiple axes.
The datasets are the following: \acrfull{boolq}, \acrfull{cb}, \acrfull{wsc}, \acrfull{axg}, \acrfull{rte}, Swedish \acrshort{cb}, and SWEDN.
Bias can be harmful and it is known to be common in data, which \acrshort{ml} models learn from.
In order to mitigate bias in data, it is crucial to be able to estimate it objectively.
We use bipol, a novel multi-axes bias metric with explainability, to estimate and explain how much bias exists in these datasets.
Multilingual, multi-axes bias evaluation is not very common.
Hence, we also contribute a new, large Swedish bias-labeled dataset (of 2 million samples), translated from the English version and train the \acrshort{sota} \acrshort{mt5} model on it.
In addition, we contribute new multi-axes lexica for bias detection in Swedish.
We make the codes, model, and new dataset publicly available.

\end{abstract}

\section{Introduction}
Recent advances in \acrfull{ai}, \acrfull{llm}, and chatbots have raised concerns about their potential risks to humanity \cite{bender2021dangers,info13060298,yudkowsky2008artificial}.\footnote{bbc.com/news/world-us-canada-65452940}
One major concern is the issue of social bias, particularly with the data \acrshort{ai} models are trained on.
Bias, which can be harmful, is the unfair prejudice in favor of or against a thing, person or group \cite{maddox2004perspectives,Dhamala2021,mehrabi2021survey,antoniak2021bad}.
Measuring bias in text data can be challenging because of the axes that may be involved (e.g. religious or gender bias).

In this work, our motivation is to determine whether social bias exists in \acrshort{nlp} benchmark datasets and estimate it.
After reviewing some potential bias methods, as discussed in Section \ref{related}, we settled for the recent bipol \cite{alkhaled2023bipol} because of its advantages.
It is a metric that estimates bias along multiple axes in text data and provides an explanation for its scores, unlike other metrics.
We investigate social bias in benchmark datasets that are available on the English SuperGLUE leaderboard and two Swedish datasets.
The SuperGLUE was introduced by \citet{NEURIPS2019_4496bf24} and provides benchmark datasets for different \acrshort{nlp} tasks.
Benchmark datasets are datasets for comparing the performance of algorithms for specific use-cases
\cite{dhar2021evaluation,paullada2021data}.
Such datasets have been the foundation for some of the significant advancements in the field \cite{paullada2021data}.
We investigate the following English datasets: 
\acrshort{boolq} \cite{clark-etal-2019-boolq}, \acrshort{cb} \cite{de2019commitmentbank}, \acrshort{wsc} \cite{levesque2012winograd}, \acrshort{axg} \cite{rudinger-etal-2018-gender}, and \acrshort{rte} \cite{NEURIPS2019_4496bf24}.
The Swedish datasets are the Overlim \textit{CB} and SWEDN.
We discuss more about the datasets in Section \ref{datasets}.


\paragraph{Our contributions}
Firstly, we show quantitatively and through explainability that bias exists in the datasets.
The findings correlate with characteristics of bias, such as heavy lopsidedness \cite{zhao-etal-2018-gender}.
This work will provide researchers with insight into how to mitigate bias in text data and possibly add impetus to the conversation on whether it is even ethical to remove these social biases from data, because they represent the real world.
Secondly, we create and release, possibly, the largest labeled dataset and lexica for bias detection in Swedish (\acrfull{mab}-Swedish) and train a model based on the \acrfull{sota} \acrfull{mt5} \cite{xue-etal-2021-mt5}.
We release our codes, dataset and artefacts publicly.
\footnote{github.com/LTU-Machine-Learning/bipolswedish.git}

The rest of this paper is structured as follows.
Section \ref{related} discusses some of the previous related work.
Section \ref{experiments} describes the methodology, including details of the characteristics of bipol and the new \acrshort{mab}-Swedish dataset.
Section \ref{results} presents the results and discusses some of the qualitative results.
In Section \ref{conclusion}, we give concluding remarks.

\section{Related Work}
\label{related}
There have been considerable effort in identifying and measuring the level of bias in datasets \cite{10.1145/3313831.3376488,Dhamala2021,stanley1977paradigmatic,chandrabose2021overview}.
These are usually targeted at gender bias in a binary form \cite{zhao-etal-2018-gender,rudinger-etal-2018-gender}.
However, studies have shown that the biases in language models for the intersection of gender and race can be greater than those for gender and race individually and that addressing bias along only one axis can lead to more issues \cite{tan2019assessing,subramanian2021evaluating}.
To determine the level of bias in \acrshort{nlp} datasets along multiple axes can be a significant challenge, more so that many of these methods admit their approaches may demonstrate the presence of bias but not prove its absence \cite{zhao-etal-2018-gender,rudinger-etal-2018-gender}.
Table \ref{table:metcompare} compares some of the methods that have been introduced.

\begin{table}[h]
\centering
\resizebox{\columnwidth}{!}{%
\begin{tabular}{lcc}
\hline
\textbf{Metric/Evaluator} & \textbf{Axis}   & \textbf{Terms}\\
\hline
WinoBias \cite{zhao-etal-2018-gender}
 & 1  &  40\\
Winogender \cite{rudinger-etal-2018-gender} & 1 &  60 \\
StereoSet \cite{nadeem-etal-2021-stereoset} & 4 & 321\\
Hurtlex \cite{nozza2021honest}
& 6 & 1,072 \\
CrowS-Pairs \citet{nangia-etal-2020-crows} & 9 & 3,016 \\
Bipol \cite{alkhaled2023bipol} & $>$2,  13*$<$ & $>$45,  466*$<$\\
 \hline
\end{tabular}
}
\caption{\label{table:metcompare} \footnotesize Comparison of some bias evaluation methods. (*The upper bounds are not limited by the bipol algorithm but the dataset \& lexica.)}
\end{table}

Furthermore, \citet{bassignana2018hurtlex} proposed a multi-language approach using HurtLex to target misogyny because addressing bias in only the English language is not sufficient for addressing the potential harm to society.
In the English language, there are common biases that associate female terms with subjects such as liberal arts and family while associating male terms with subjects such as science \cite{nosek2002harvesting}.
There are also more words that sexualize females more than males \cite{stanley1977paradigmatic}. Other languages have their own peculiarities \cite{nozza2021honest}.

In addition to the various methods identified in Table \ref{table:metcompare} for quantifying the extent of discrimination or bias, there is also odds ratio (OR), which compares the chance of a specific outcome happening, with a certain exposure, to the likelihood of that outcome happening without the exposure \cite{szumilas2010explaining}.
Another method is the impact ratio (IR), which calculates the ratio of positive outcomes for a protected group to the general group.
In \citet{10.1145/3313831.3376488}, they compare lexicon method to model classification for gender bias in English language only.
Our approach combines the strengths of both approaches and evaluates on English and Swedish data across multiple axes.


\section{Methodology}
\label{experiments}

\subsection{Bipol}
\label{bipol}
There are two stages in the implementation of bipol (see \ref{eq:eq1}) before it gives a final score between 0.0 (zero or undetected bias) and 1.0 (extreme bias).
The first stage involves the classification of the data samples (into biased and unbiased categories) using a trained model (see \ref{eq:eq2}).
Ideally, it is the ratio of the number of  \acrfull{tp} to the total samples (\acrfull{tp}, \acrfull{fp}, \acrfull{tn}, and \acrfull{fn}), where \acrshort{fp} is preferably zero.
However, since the trained models will be evaluated on unseen data, the predicted biased samples are likely to have \acrshort{fp} in the numerator as expressed in the equation. 
The evaluations thus come with positive error rate (\( \frac{\acrshort{fp}}{\acrshort{fp} + \acrshort{tp}} \)) to establish the lower bound of error for the predictions.
A good classifier should minimize the number of \acrshort{fp} and maximize the number of \acrshort{tp}  but there's hardly any perfect classifier, even in other tasks such as spam detection or hate speech \citep{heron2009technologies,feng2018multistage}.

\begin{subequations}
\small

\begin{equation}
\mathit{b}=\begin{cases}
\mathit{b_{c}} . \mathit{b_{s}}, & \text{if $b_{s}>0$}\\
\mathit{b_{c}}, & \text{otherwise}
\end{cases}
\label{eq:eq1}
\end{equation}

\begin{equation}
\mathit{b_{c}} = 
\frac{tp + fp}{tp+fp+tn+fn}
\label{eq:eq2}
\end{equation}

\begin{equation}
\mathit{b_{s}} = \frac{1}{r} \sum_{t=1}^{r} 
{\left( \frac{1}{q} \sum_{x=1}^{q} {\left(
\frac{|\sum_{s=1}^{n} a_{s} - \sum_{s=1}^{m} c_{s}|}{\sum_{s=1}^{p} d_{s}}
\right)}_{x}
\right)}_{t}
\label{eq:eq3}
\end{equation}
\end{subequations}

The second stage evaluates the biased samples for sensitive terms listed in the multi-axes lexica (see \ref{eq:eq3}). 
It involves finding the difference between the two maximum summed frequencies in the types (e.g. female) of an axis (e.g. gender) (\(|\sum_{s=1}^{n} a_{s} - \sum_{s=1}^{m} c_{s}| \)), which is then divided by the summed frequencies of all the terms in that axis (\( \sum_{s=1}^{p} d_{s} \)).
The average over all the axes (\( \frac{1}{q} \sum_{x=1}^{q} \)) is then averaged over all the biased samples (\( \frac{1}{r} \sum_{t=1}^{r} \) ).
Table \ref{table:lexi} provides the Swedish lexica sizes.
The lexica are derived from \citet{adewumi2020corpora,adewumi2020exploring} and Wikipedia\footnote{en.wikipedia.org/wiki/Swedish\_profanity} and may be expanded as needed.
They include terms that may be stereotypically associated with certain groups and specific gender \cite{10.1145/3313831.3376488,zhao-etal-2018-gender}.
The English lexica contain more and are also derived from  public sources \cite{alkhaled2023bipol}.

\begin{table}[h]
\small
\centering
\begin{tabular}{lcc}
\hline
\textbf{Axis} & \textbf{Axis type 1} & \textbf{Axis type 2} \\
\hline
Gender & 17 (female) & 19 (male)\\
Racial & 10 (black) & 10 (white)\\
\hline
\end{tabular}
\caption{\label{table:lexi}
Swedish lexica sizes. These may be expanded.
}
\end{table}

The rationale for using bipol is because of the strengths of the metric.
These include 1) the relative simplicity of calculating a score, 2) it is straight-forward to implement since it is based on existing concepts like lexica and classifiers, 3) it captures semantic and \acrfull{tf} aspects of data, 4) it has explainability built in, 5) it's possible to determine the error rate of predictions, and 6) it is not limited in the total number of axes that may be used.
We acknowledge, however, that it has limitations that are based on the limitations of the tools that may be used to calculate it.

\subsection{Datasets}
\label{datasets}
\subsubsection{The New \acrshort{mab}-Swedish Dataset}
The dataset was machine-translated (from \acrshort{mab} \cite{alkhaled2023bipol}) using the Helsinki-\acrshort{nlp} model \cite{TiedemannThottingal:EAMT2020}, which was mostly trained with guided alignment.
The automatic translation took over 48 hours on one GPU.
It has 1,946,975 samples, as given in Table \ref{table:genbias}.
Quality control (QC) for the \acrshort{mab}-Swedish involved translation verification by back-translating some random samples using Google \acrshort{nmt} before a review by 
 a Swedish speaker.
The English version was constructed from two datasets: Jigsaw\footnote{medium.com/jigsaw/creating-labeled-datasets-and-exploring-the-role-of-human-raters-56367b6db298} and the \acrfull{sbic} by \cite{sap-etal-2020-social}.
\acrfull{pii} were removed from the dataset.
More details about the annotation of the base datasets for the \acrshort{mab} can be found in \citet{alkhaled2023bipol}.
Some examples in the \acrshort{mab}-Swedish are given in Table \ref{table:mabsamples}.

\paragraph{Machine-Translation concerns}
Bias is a universal concern, though there can be culture-specific biases.
A stereotype or degrading comment can be considered of universal concern if it is relevant across cultures.
For instance, the first example in Table \ref{table:mabsamples} is of universal concern because women should be respected, regardless of nationality or culture.
This also applies to the other biased examples in the table.
Hence, \acrshort{mab}-Swedish is relevant for bias detection in Swedish, though it was translated.
A limitation of the dataset, however, is that it may not contain examples that are specific to the Swedish culture since the original dataset is from English, with specific references to the USA and Canada.

\begin{table}[h]
\small
\centering
\begin{tabular}{lccc}
\hline
\textbf{Set} & \textbf{Biased} & \textbf{Unbiased} & \textbf{Total}\\
\hline
Training & 533,544 & 1,209,433 & 1,742,977 \\
Validation & 32,338 & 69,649 & 101,987\\
Test & 33,470 & 68,541 & 102,011\\
\hline
 & 599,352 & 1,347,623 & 1,946,975 \\
 \hline
\end{tabular}
\caption{\label{table:genbias}
 \acrshort{mab}-Swedish dataset split.
}
\end{table}

\begin{table}[h]
\small
\centering
\begin{tabular}{p{0.45\linewidth} | p{0.12\linewidth} | 
 p{0.1\linewidth} | p{0.1\linewidth}}
\hline
\textbf{comment\_text} & \textbf{label} & \textbf{old\_id} & \textbf{id}  
\\
\hline
Den här slynan är galen som skulle läsa en bok av en kvinna. & biased & 239612 & 1212584 \\
(\textit{This b*tch is nuts. Who would read a book by a woman.}) & & & \\
 9 av 10 människor njuter av mobbning. Jag kommer alltid att vara den 10: e & biased & none & 1517517\\
(\textit{9 out of 10 people enjoy bullying. I will always be the 10th.}) & & & \\
"Författaren säger att det finns > många, många fler anledningar till att jag antar att en uppföljare finns i verken, då?" & unbiased & 383811 & 110831\\
(\textit{"The author says that there are many, many more reasons I am assuming a sequel is in the works, then?"}) & & & \\
Vad kallar du underkläder för araber?& biased & none & 1618146 \\
(\textit{What do you call lingerie for Arabs? Socks.}) & & & \\
\hline
\end{tabular}
\caption{\label{table:mabsamples}
Examples from the \acrshort{mab}-Swedish (The English in the original is in italics.)
}
\end{table}

\subsubsection{\acrshort{boolq}}
It is a \acrfull{qa} task where each example has a short passage and a yes/no question about the passage \cite{clark-etal-2019-boolq} .
These questions were provided anonymously by Google search users and afterwards paired with a paragraph from a Wikipedia article that has the answer.
We evaluated the passage column of the dataset.

\subsubsection{\acrshort{cb}} This contains short texts in which, at least, one sentence has an embedded clause \cite{de2019commitmentbank}.
The resulting task is framed as three-class textual entailment on examples that are drawn from the following datasets: Wall Street Journal, fiction from the British National Corpus, and Switchboard.
We evaluated the premise column of the dataset.

\subsubsection{\acrshort{wsc}} This is a coreference resolution dataset \cite{levesque2012winograd}.
Examples consist of a sentence with a pronoun and a list of noun phrases from the sentence.
We evaluated the text column of the dataset.

\subsubsection{\acrshort{axg}} It is designed to measure gender bias in coreference resolution systems \cite{rudinger-EtAl:2018:N18}.
Each example consists of a premise sentence having a male or female pronoun and a hypothesis giving a possible antecedent of the pronoun.
We evaluated the premise column of the dataset.

\subsubsection{\acrshort{rte}} The datasets come from a series of annual competitions on textual
entailment \cite{NEURIPS2019_4496bf24}.
Data from several sources were merged and converted to two-class classification:
entailment and not\_entailment.
We evaluated the premise column of the dataset.

\subsubsection{Swedish \acrshort{cb}}
This is part of the OverLim dataset by the National Library of Sweden.
It contains some of the GLUE and SuperGLUE tasks automatically translated to Swedish, Danish, and Norwegian, using the OpusMT models for MarianMT\footnote{huggingface.co/datasets/KBLab/overlim}.
We evaluated its training set.

\subsubsection{SWEDN}
This is a text summarization corpus based on 1,963,576 news articles from the Swedish newspaper Dagens Nyheter (DN) during the years 2000 to 2020.\footnote{spraakbanken.gu.se/resurser/swedn}
There are five categories of articles in the dataset: domestic news, economy, sports, culture, and others \cite{monsen2021method}.
The training set consists of the first three categories and we evaluate the first 1,000 samples because of the computation cost of evaluation.

\subsection{Experiments}

The experiments were conducted on two shared Nvidia DGX-1 clusters running Ubuntu 18.04 and 20.04 with 8 × 32GB V100 and 8 x 40GB A100 GPUs, respectively.
Average results are reported after running each experiment twice.
To evaluate the benchmark datasets, we utilize bias-detection models \cite{alkhaled2023bipol} based on \acrshort{roberta} \citep{liu2019roberta}, Electra \citep{clark2020electra}, and DeBERTa \citep{he2021deberta}.
We train a small \acrshort{mt5} model with batch size of 16, due to memory constraints, on the \acrshort{mab}-Swedish.
Wandb \cite{wandb}, an experiment tracking tool, is run for 5 counts with bayesian optimization to suggest the best hyper-parameter combination for the learning rate (1e-3 - 2e-5) and epochs (6 - 10) before final training of the model.
We use the pretrained model from the HuggingFace hub \cite{wolf-etal-2020-transformers}.
Average training time was 15 hours.
Average evaluation time ranges from about 30 minutes to over 24 hours.\footnote{particularly when cpulimit is used, in fairness to other users}

\section{Results and Discussion}
\label{results}
From Table \ref{table:res} we observe that all the datasets have bias, though little, given that they are smaller than a \textit{bipol} score of 1.
The dataset with the least amount of bias is \acrshort{boolq}, which is confirmed by all the three models. 
This is despite the dataset having the highest number of unique samples.
\acrshort{cb} has the largest amount of bias and this is also confirmed by the three models.
This is also the case for the Swedish \acrshort{cb}, when compared with SWEDN.

The average macro F1 score on the validation set of \acrshort{mab}-Swedish is 0.7623 with standard deviation (s.d.) of 0.0075.
The resulting error rate is 0.2893.
This is relatively reasonable though a bit higher than the error rate for the English \acrshort{roberta}, Electra, and DeBERTa, which are 0.198, 0.196, and 0.2, respectively \cite{alkhaled2023bipol}.

\begin{table}[h]
\small
\centering
\begin{tabular}{ccccc}
\hline
 & & \multicolumn{3}{c}{\textbf{bipol level} $\downarrow$ (s.d.)} \\
\acrshort{roberta} & samples & \textbf{corpus} & \textbf{sentence} & \textbf{bipol \textit{(b)}} \\
\hline
\acrshort{boolq} & 7,929 & 0.0066 & 0.8027 & 0.0053 (0)\\
\acrshort{cb} & 250 & 0.08 & 0.8483 & 0.0679 (0)\\
\acrshort{wsc} & 279 & 0.0466 & 0.8718 & 0.0406 (0)\\
\acrshort{axg} & 178 & 0.0112 & 1 & 0.0112 (0)\\
\acrshort{rte} & 2,379 & 0.0294 & 0.8518 & 0.0251 (0)\\
 \\ \hline

 Electra & & & & \\
\hline
\acrshort{boolq} & 7,929 & 0.0073 & 0.8089 & 0.0059 (0)\\
\acrshort{cb} & 250 & 0.0316 & 0.881 & 0.074 (0)\\
\acrshort{wsc} & 279 & 0.0609 & 0.9559 & 0.0582 (0)\\
\acrshort{axg} & 178 & 0.0112 & 1 & 0.0112 (0) \\
\acrshort{rte} & 2,379 & 0.0269 & 0.8593 & 0.0231 (0)\\
 \\ \hline

DeBERTa & & & & \\
\hline
\acrshort{boolq} & 7,929 & 0.0103 & 0.7212 & 0.0075 (0) \\
\acrshort{cb} & 250 & 0.084 & 0.9048 & 0.076 (0) \\
\acrshort{wsc} & 279 & 0.0609 & 1 & 0.0609 (0) \\
\acrshort{axg} & 178 & 0.0112 & 1 & 0.0112 (0) \\
\acrshort{rte} & 2,379 & 0.0366 & 0.8655 & 0.0316 (0)\\
 \\ \hline
 
\multicolumn{5}{l}{mT5 on Swedish data} \\
\hline
\acrshort{cb} & 201 & 0.0796 & 0.7188 & 0.0572 (0)\\
SWEDN & 1,000 & 0.053 & 0.9433 & 0.05 (0)\\
 \\ \hline

\end{tabular}
\caption{\label{table:res}
Results of average bipol scores. All the datasets have bias, though little.
}
\end{table}

\subsection{Error Analysis}
\begin{figure}[h!]
\centering
\includegraphics[width=0.5\textwidth]{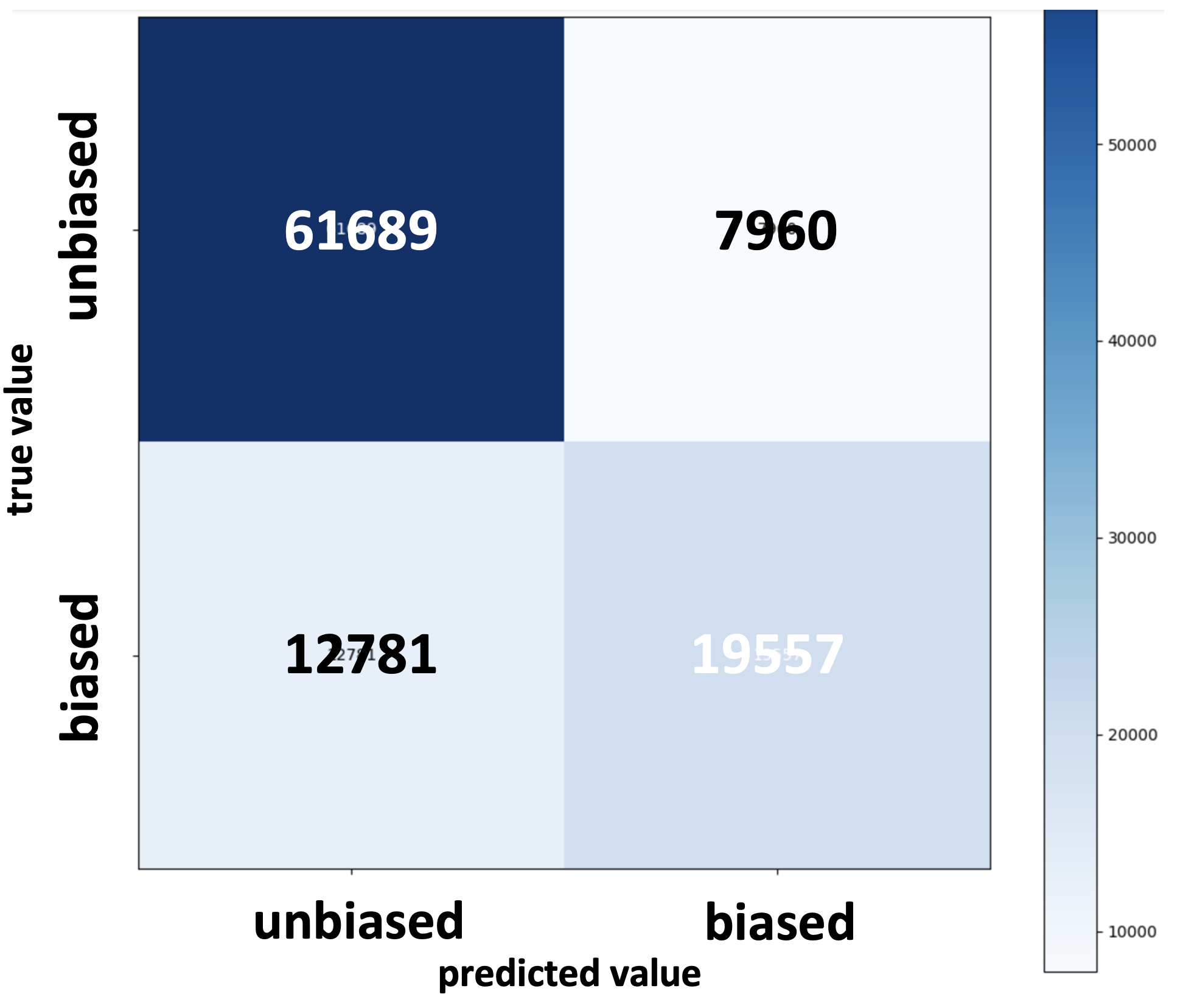}
\caption{Confusion matrix for \acrshort{mt5}.}
\label{cmatrix}
\end{figure}

Figure \ref{cmatrix} presents the confusion matrix for the \acrshort{mt5} on the \acrshort{mab}-Swedish.
The tn, fp, fn, tp are 61,689, 7,960, 12,781, and 19,557, respectively, which are relevant for Eq. \ref{eq:eq2}.
We observe that the model is better at predicting unbiased samples.
This is expected since the training data contains more examples of unbiased samples. 
Table \ref{table:qexamples} presents some qualitative examples of apparently correct and incorrect predictions in two of the datasets.
The first correct example in the English \acrshort{cb} appears to have a clear stereotype that \textit{men are naturally right and it is the role of women to follow their lead}.
The second correct example, in both the English and Swedish data, may have been perceived as biased by the two different models because of the offensive term \textit{fool} or the overgeneralization that \textit{folk will always take advantage
of weakness} or both.
Overgeneralization is a characteristic of bias \cite{rudinger-etal-2018-gender,nadeem-etal-2021-stereoset}.

\begin{table*}[h!]
\small
\centering
\begin{tabular}{p{0.58\linewidth} | p{0.12\linewidth} | p{0.1\linewidth}}
\hline
\textbf{examples} & \textbf{label} & \textbf{csv id}  
\\
\hline
\multicolumn{3}{c}{\textbf{Correct predictions from the English \acrshort{cb}, based on DeBERTa}}
\\
\hline
nicky approached her with the assumption that men are naturally right and it is the role of women to follow their lead. constance whose confidence was growing daily was not prepared to give in to nickys wishes merely because of his sex. if she felt he was right then she agreed with him. & biased & 85 \\
\hline
she rubbed them away with an angry fist. she was a fool to let anyone get round her. how long before she learned that folk ll always take advantage of weakness & biased & 73\\
\hline
\multicolumn{3}{c}{\textbf{Incorrect prediction}}
\\
\hline
he d gone. philip had to get them back. his dad would kill him if he found that he d taken them.& biased & 107
\\
\hline
& & 
\\
\hline
\multicolumn{3}{c}{\textbf{Correct predictions from the Swedish \acrshort{cb}, based on m\acrshort{t5}}}
\\
\hline
det är en del av deras religion en religion som jag inte hånar eftersom den innehåller många element som matchar vår egen trots att den saknar vår sanning. vid en av deras stora högtider har de ritualen att driva ut djävlarna från sina kroppar. först kommer trummisarna jag kan säga att inga kvinnor tillåts delta i denna ritual och damerna här kanske håller med mig om att de är lyckligt lottade i detta utelämnande. & biased & 2
\\
(\textit{it is part of their religion a religion that I do not mock because it contains many elements that match our own even though it lacks our truth. at one of their great festivals they have the ritual of casting the devils out of their bodies. first come the drummers i can say no women are allowed to participate in this ritual and the ladies here may agree with me that they are lucky in this omission.})
&  & \\
\hline
hon gnuggade bort dem med en arg knytnäve. hon var en idiot som lät vem som helst komma runt henne. hur lång tid tar det innan hon får veta att folk alltid kommer att utnyttja svaghet & biased & 59 \\
(\textit{she rubbed them away with an angry fist. she was a fool to let anyone get round her. how long before she learned that folk will always take advantage of weakness}) &  &  \\
\hline
\multicolumn{3}{c}{\textbf{Incorrect prediction}}
\\
\hline
han var borta. philip var tvungen att få tillbaka dem. hans pappa skulle döda honom om han fann att han hade tagit dem.& biased & 86 \\
(\textit{he had gone. philip had to get them back. his dad would kill him if he found that he had taken them.})
& & 
\\
\hline
\end{tabular}
\caption{\label{table:qexamples}
Qualitative examples of apparently correct and incorrect predictions in some of the datasets. The English translations appear in \textit{italics} for the Swedish examples.
}
\end{table*}


\subsection*{Explaining bias type}
\begin{figure}[h!]
\centering
\includegraphics[width=0.5\textwidth]{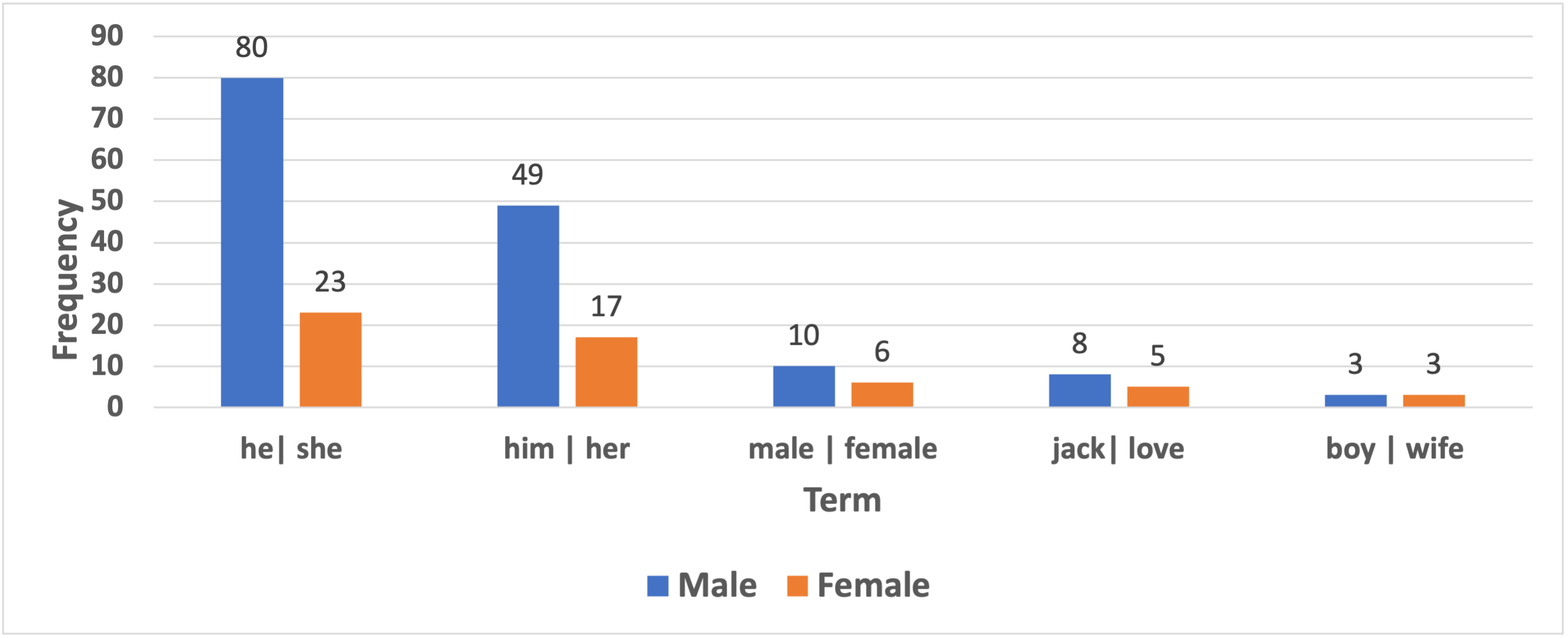}
\caption{Top-5 gender frequent terms in \acrshort{boolq} by \acrshort{roberta}.}
\label{ro_boolq}
\end{figure}

\begin{figure}[h!]
\centering
\includegraphics[width=0.5\textwidth]{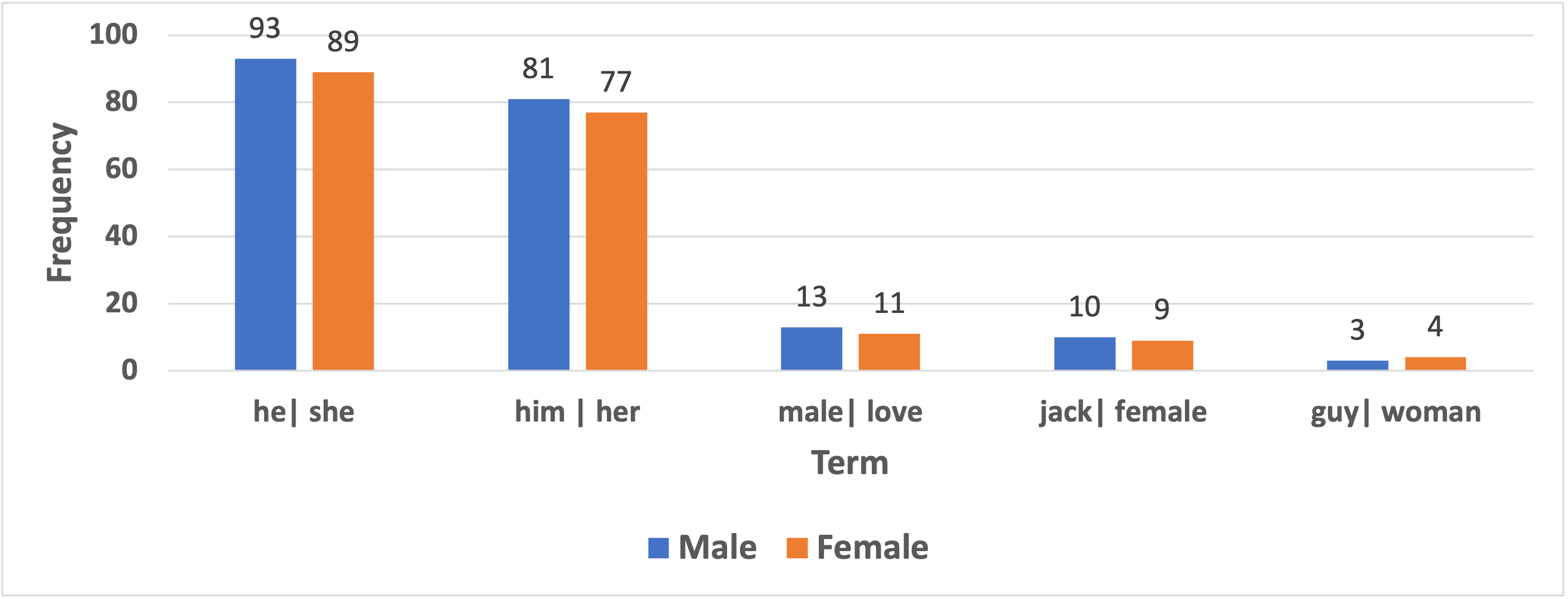}
\caption{Top-5 gender frequent terms in \acrshort{boolq} by DeBERTa.}
\label{de_boolq}
\end{figure}

\begin{figure}[h!]
\centering
\includegraphics[width=0.5\textwidth]{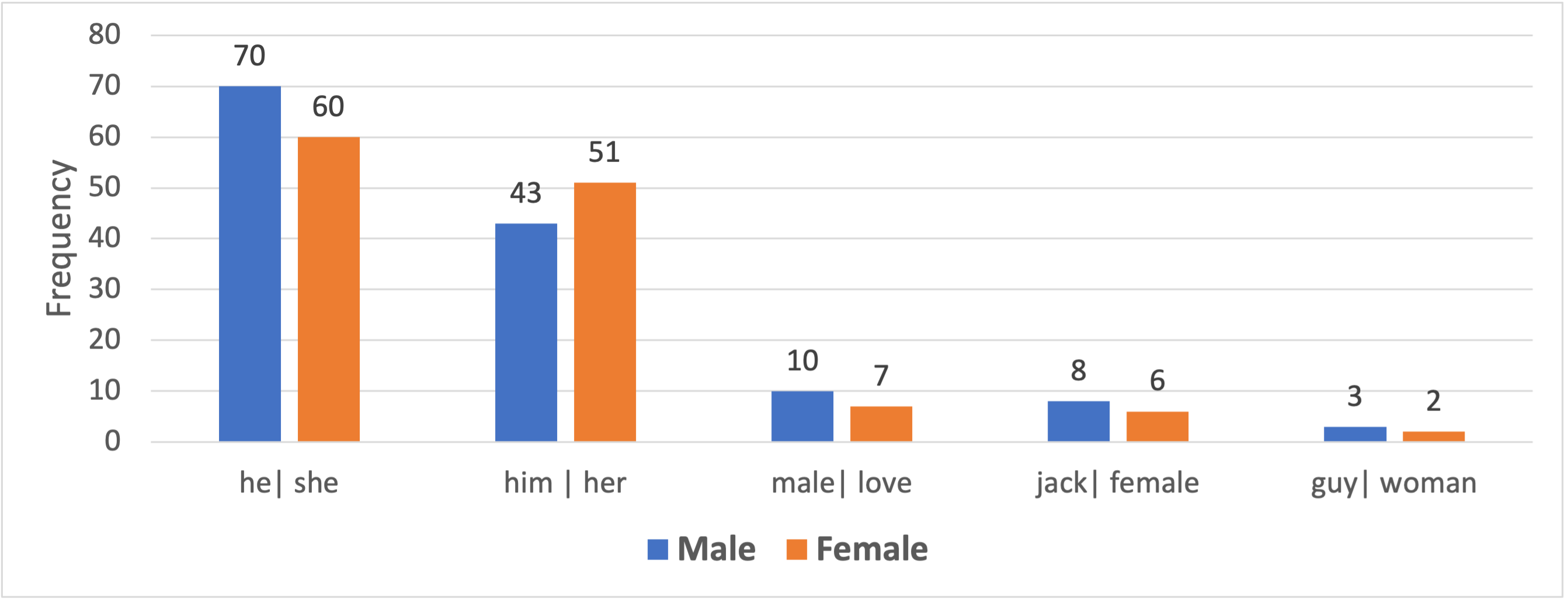}
\caption{Top-5 gender frequent terms in \acrshort{boolq} by Electra.}
\label{el_boolq}
\end{figure}

\begin{figure}[h!]
\centering
\includegraphics[width=0.5\textwidth]{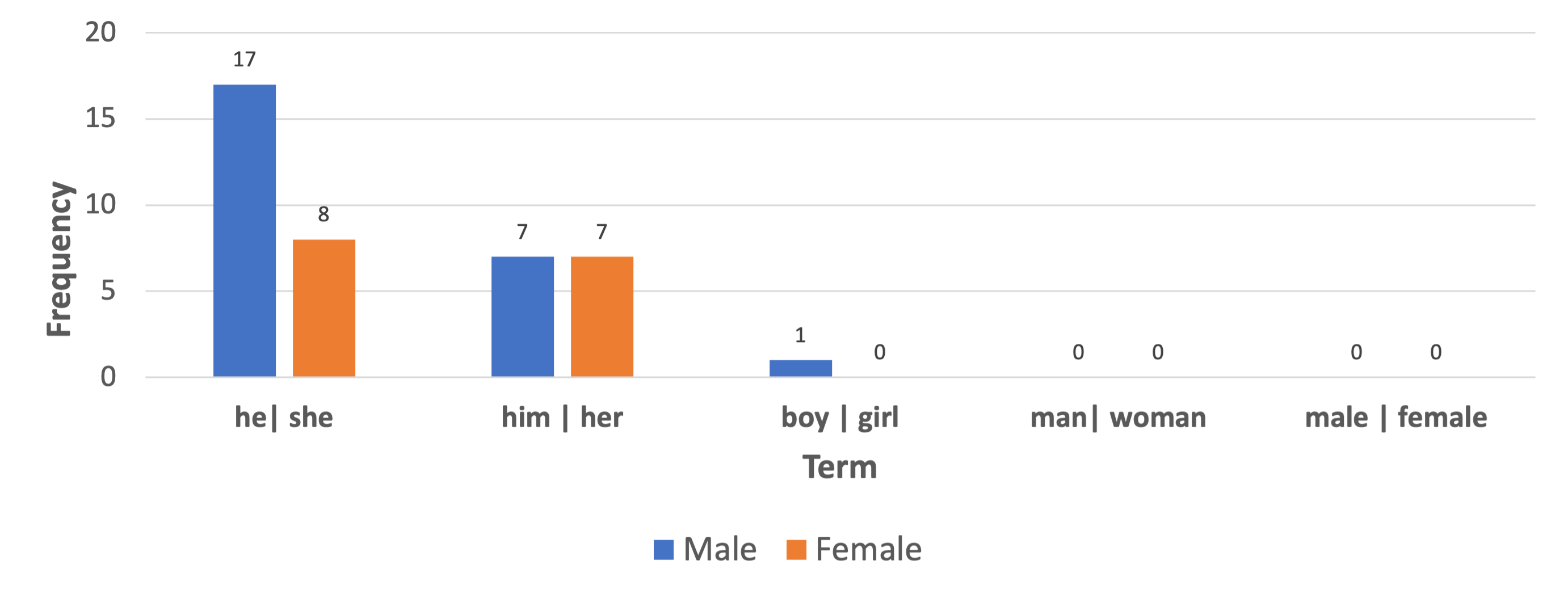}
\caption{Top-5 gender frequent terms in \acrshort{cb} by Roberta.}
\label{ro_cb}
\end{figure}

\begin{figure}[h!]
\centering
\includegraphics[width=0.5\textwidth]{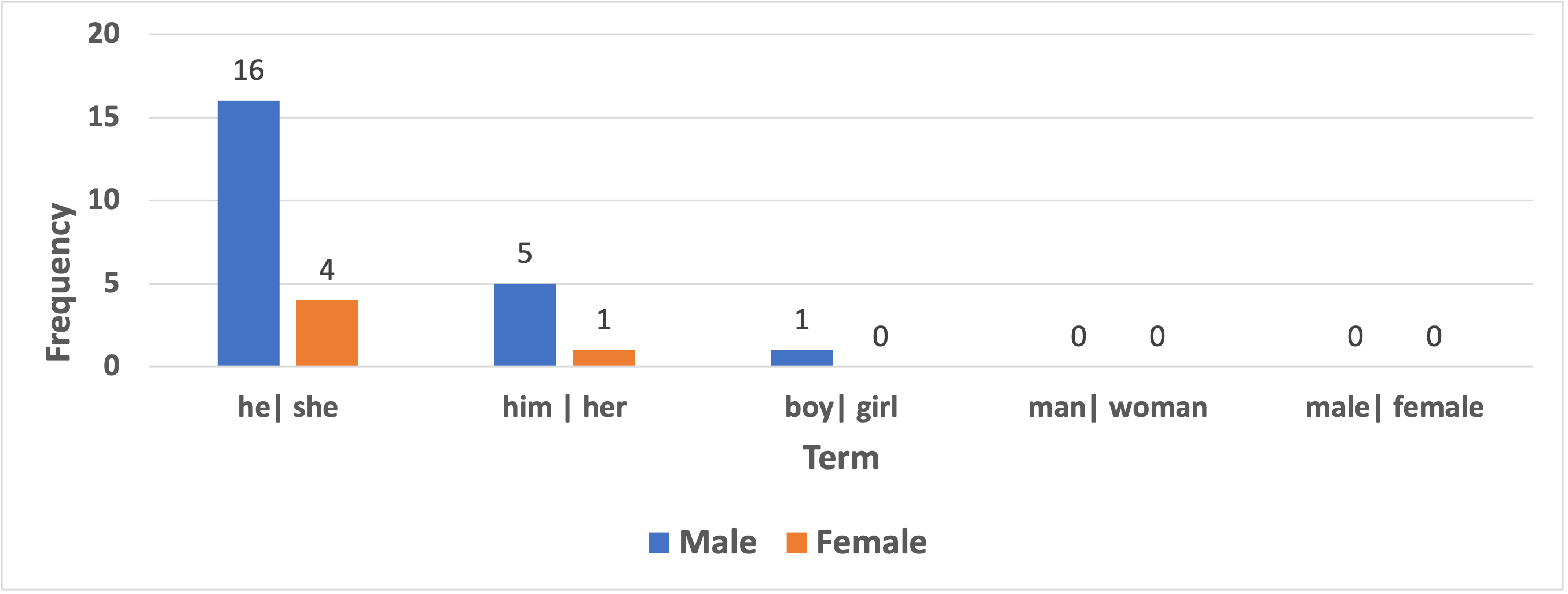}
\caption{Top-5 gender frequent terms in \acrshort{cb} by DeBERTa.}
\label{de_cb}
\end{figure}

\begin{figure}[h!]
\centering
\includegraphics[width=0.5\textwidth]{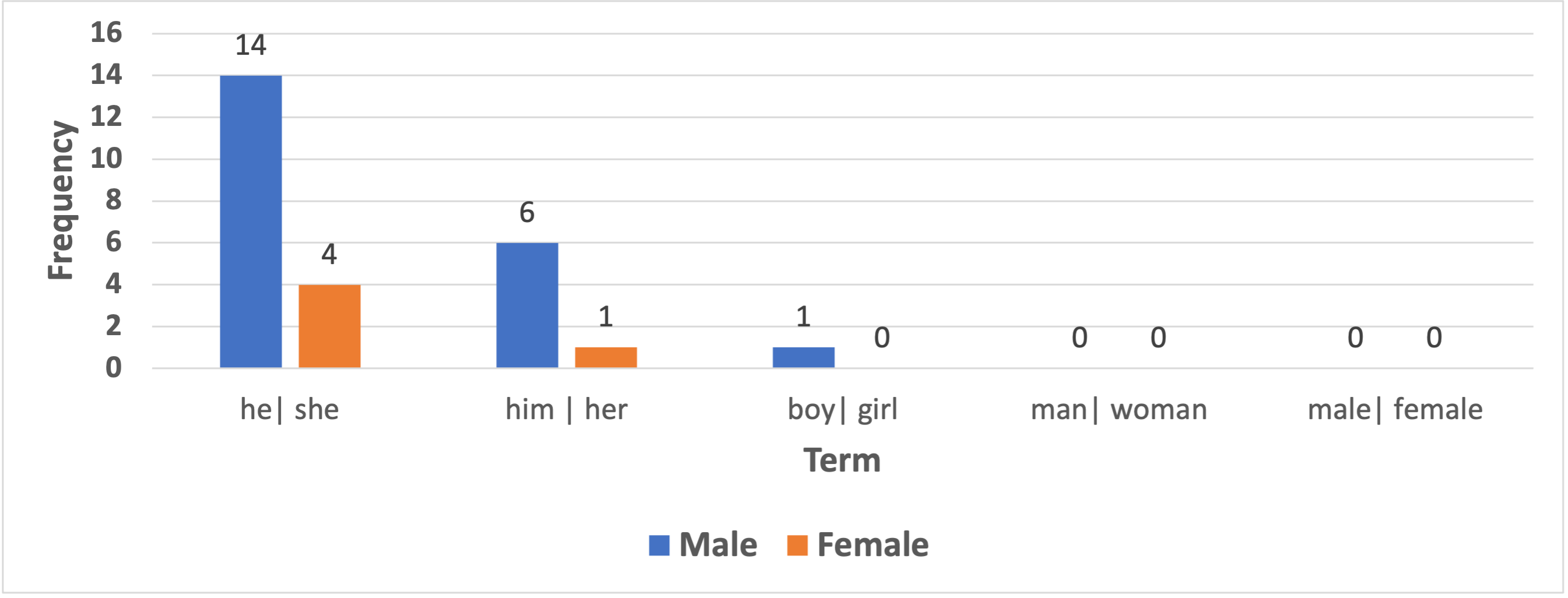}
\caption{Top-5 gender frequent terms in \acrshort{cb} by Electra.}
\label{el_cb}
\end{figure}

\begin{figure}[h!]
\centering
\includegraphics[width=0.5\textwidth]{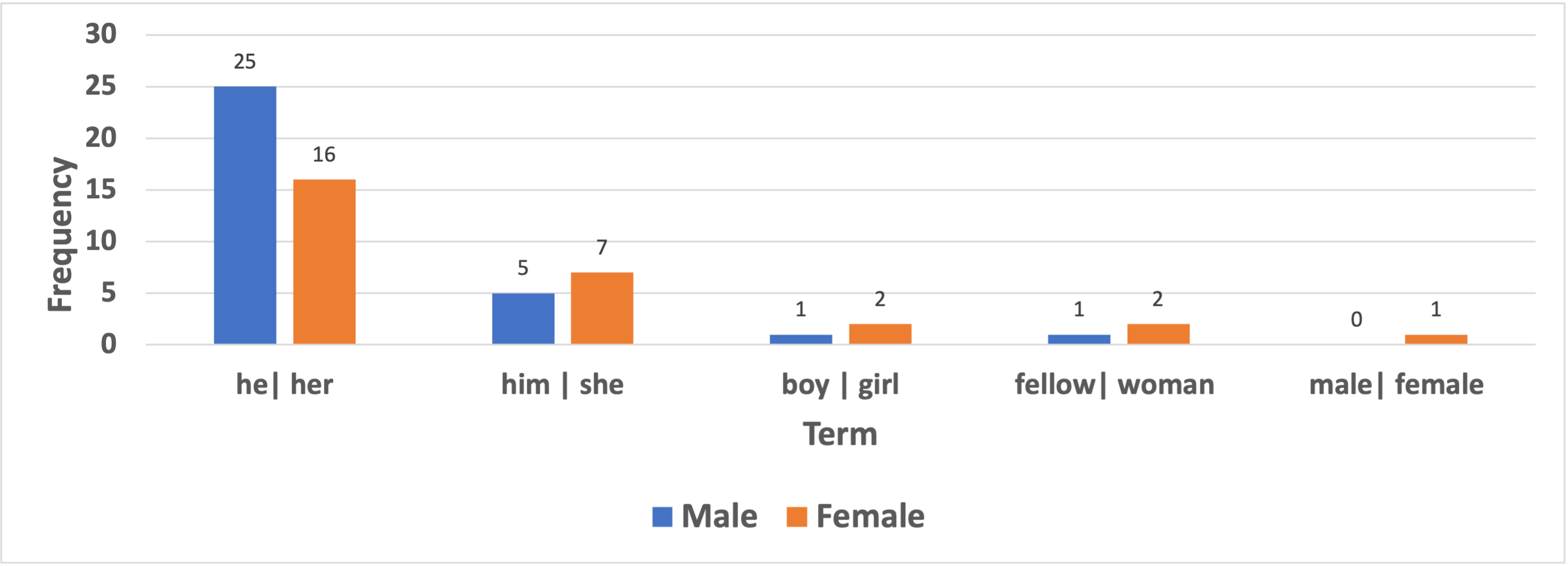}
\caption{Top-5 gender frequent terms in \acrshort{rte} by RoBERTa.}
\label{ro_rte}
\end{figure}

\begin{figure}[h!]
\centering
\includegraphics[width=0.5\textwidth]{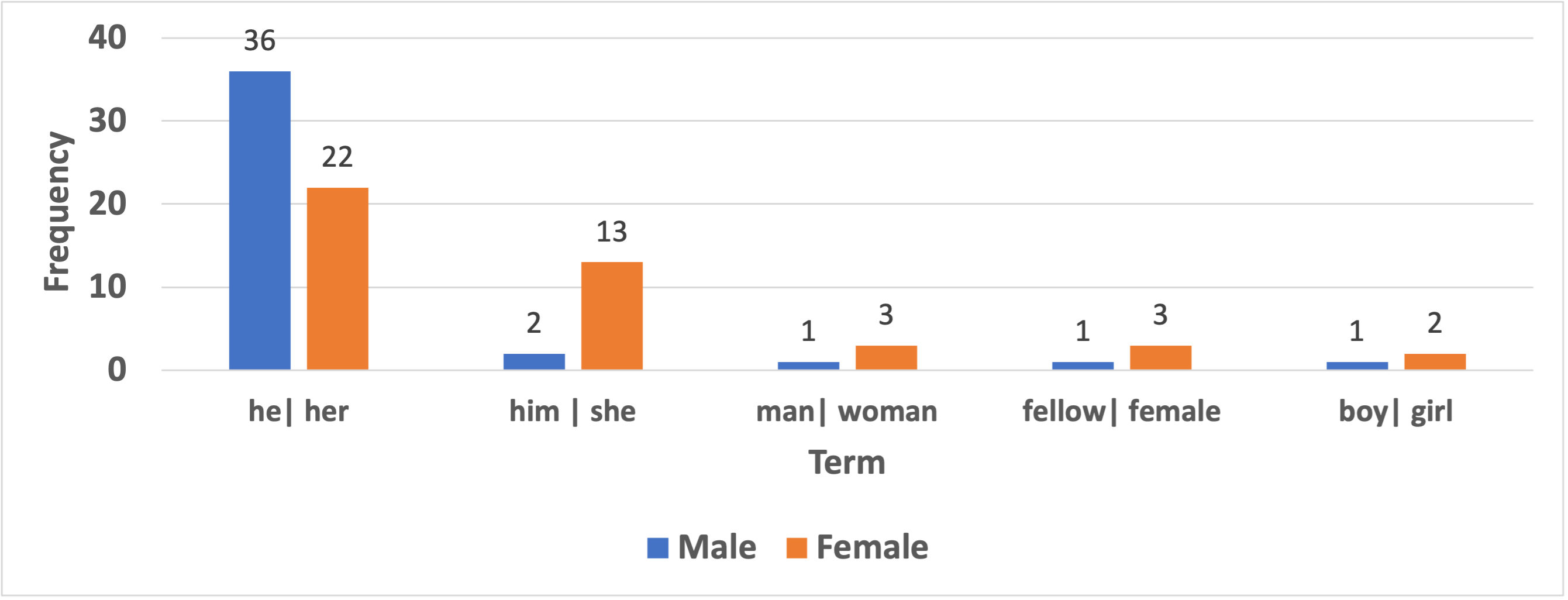}
\caption{Top-5 gender frequent terms in \acrshort{rte} by DeBERTa.}
\label{de_rte}
\end{figure}

\begin{figure}[h!]
\centering
\includegraphics[width=0.5\textwidth]{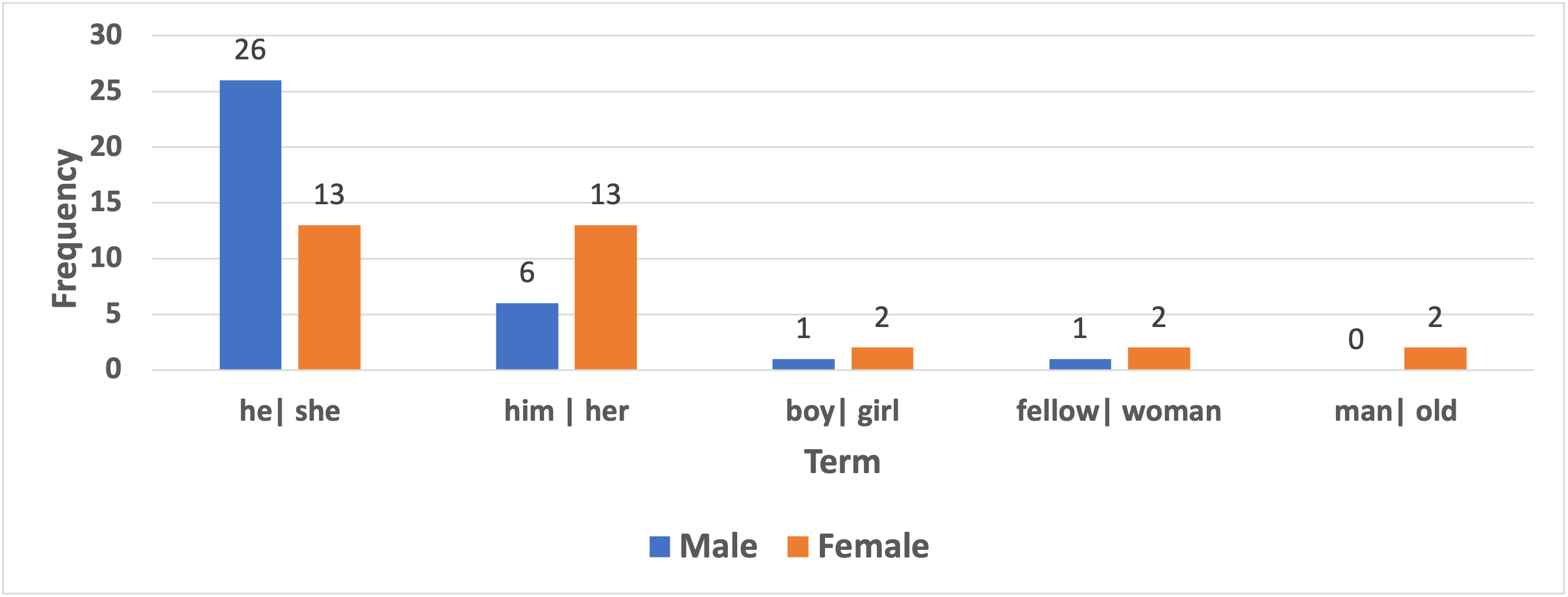}
\caption{Top-5 gender frequent terms in \acrshort{rte} by Electra.}
\label{el_rte}
\end{figure}

\begin{figure}[h!]
\centering
\includegraphics[width=0.5\textwidth]{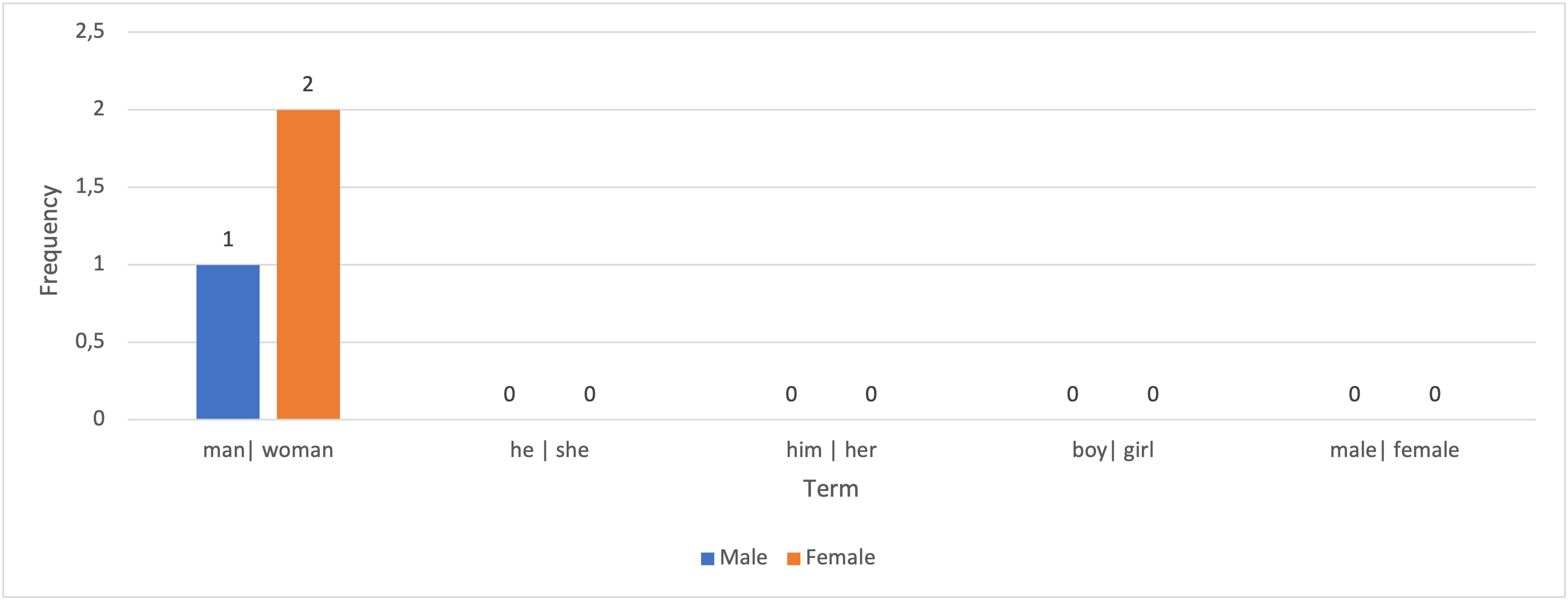}
\caption{Top-5 gender frequent terms in \acrshort{wsc} by RoBERTa.}
\label{ro_wsc}
\end{figure}

\begin{figure}[h!]
\centering
\includegraphics[width=0.5\textwidth]{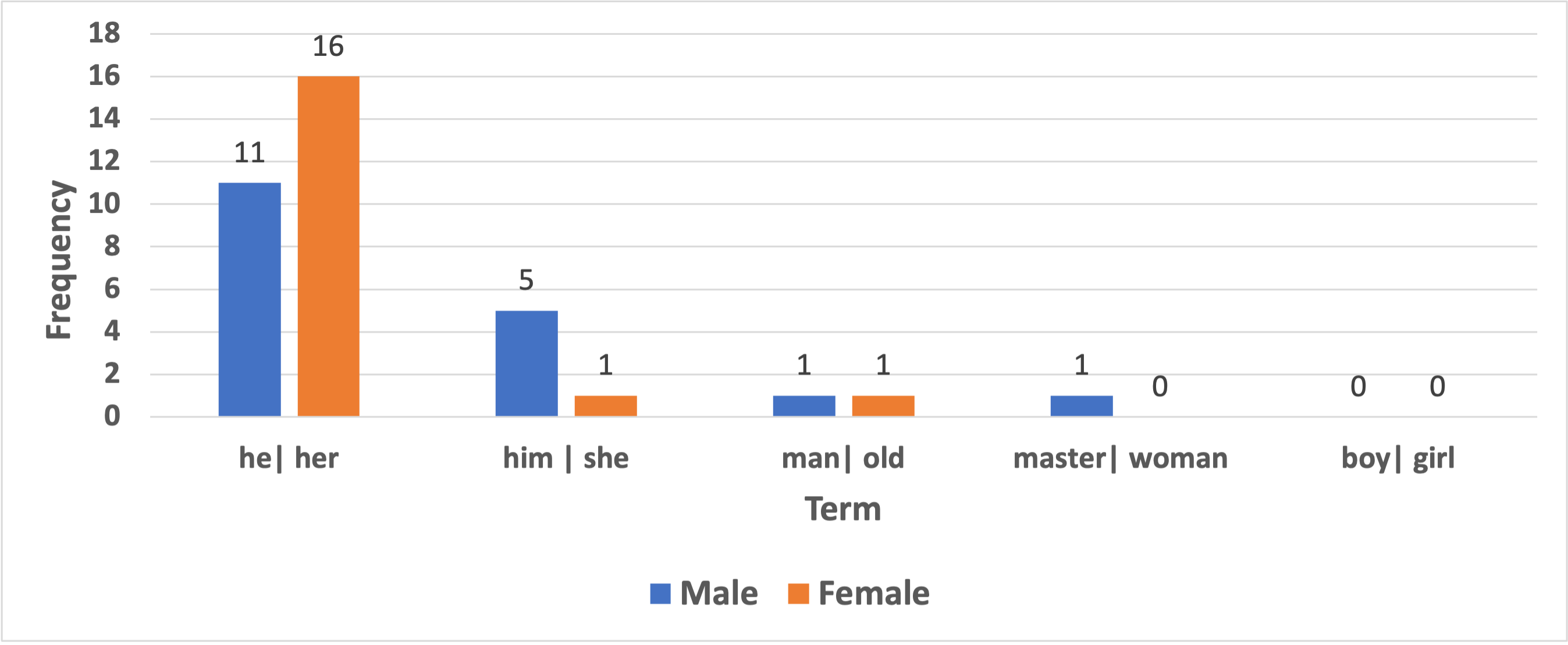}
\caption{Top-5 gender frequent terms in \acrshort{wsc} by DeBERTa.}
\label{de_wsc}
\end{figure}

\begin{figure}[h!]
\centering
\includegraphics[width=0.5\textwidth]{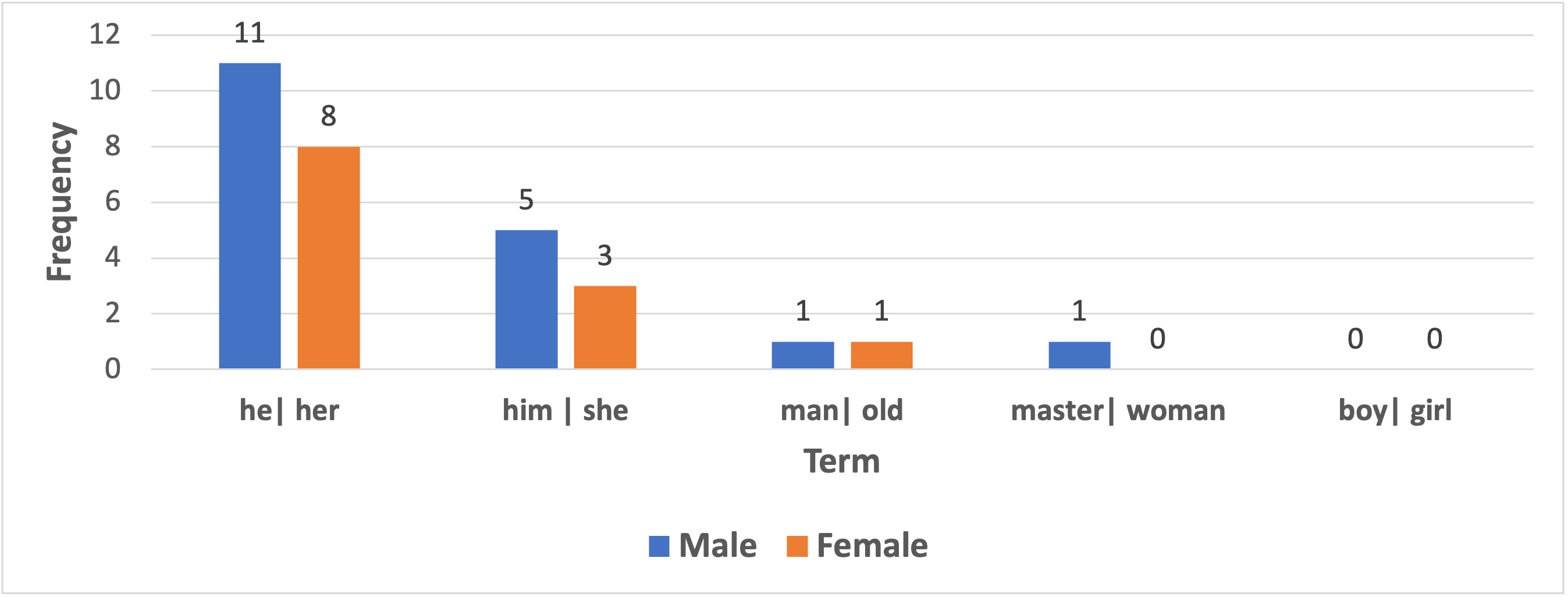}
\caption{Top-5 gender frequent terms in \acrshort{wsc} by Electra.}
\label{el_wsc}
\end{figure}

The type of overall bias (for the gender axis) in many of the datasets is explained by the dictionary of lists produced by bipol (see Appendix \ref{experiment}) and represented in "top-5 frequent terms" bar graphs
of Figures \ref{ro_boolq} to \ref{el_wsc}.
As expected, we observed that \acrshort{axg} is limited to only gender, unlike \acrshort{boolq}, which also reflects religious bias, as explained in their bipol dictionaries of lists.
We observe from Figures \ref{ro_boolq}, \ref{de_boolq}, and \ref{el_boolq} that \acrshort{boolq} is male-biased.
Figures \ref{ro_cb}, \ref{de_cb}, and \ref{el_cb} show that \acrshort{cb} is also male-biased.
This is the case also for \acrshort{rte}, as revealed by Figures \ref{ro_rte}, \ref{de_rte}, and \ref{el_rte}.
On the other hand, we observe that the case of \acrshort{wsc} is not clear-cut because Figure \ref{ro_wsc} shows only a marginal lead for female bias, Figure \ref{de_wsc} shows the difference among the top-5 is zero and Figure \ref{el_wsc} shows a slight overall male bias.

\section{Conclusion}
\label{conclusion}
We show that all benchmark datasets we evaluated, including the Swedish datasets, contain bias to different degrees.
This is likely the first time these datasets are evaluated in such a way that estimates the amount of bias and the type.
We believe these evaluations will motivate research on how to more effectively mitigate bias along multiple axes in datasets.
This work may encourage discussions on whether the biased samples from the benchmark datasets should be disregarded entirely or if they should be utilized in a different manner than previously done.
Our public release of the new \acrshort{mab}-Swedish dataset, lexica and model will also facilitate future work in multilingual bias detection.

\section*{Ethics Statement}
The authors made the effort to obscure offensive terms in examples that were used in this paper. 
We note that the models for estimating the biases in the datasets are limited in scope, as they  only cover certain number of axes (12).
Therefore, a result of 0 on any dataset does not necessarily indicate a bias-free dataset.

\section*{Acknowledgments}
We sincerely thank the anonymous reviewers for their valuable feedback on this work.

\bibliographystyle{acl_natbib}
\bibliography{ranlp2023}

\begin{thebibliography}{41}
\expandafter\ifx\csname natexlab\endcsname\relax\def\natexlab#1{#1}\fi

\bibitem[{Adewumi et~al.(2022)Adewumi, Liwicki, and Liwicki}]{info13060298}
Tosin Adewumi, Foteini Liwicki, and Marcus Liwicki. 2022.
\newblock \href {https://doi.org/10.3390/info13060298} {State-of-the-art in
  open-domain conversational ai: A survey}.
\newblock \emph{Information}, 13(6).

\bibitem[{Adewumi et~al.(2020{\natexlab{a}})Adewumi, Liwicki, and
  Liwicki}]{adewumi2020corpora}
Tosin~P Adewumi, Foteini Liwicki, and Marcus Liwicki. 2020{\natexlab{a}}.
\newblock Corpora compared: The case of the swedish gigaword \& wikipedia
  corpora.
\newblock \emph{arXiv preprint arXiv:2011.03281}.

\bibitem[{Adewumi et~al.(2020{\natexlab{b}})Adewumi, Liwicki, and
  Liwicki}]{adewumi2020exploring}
Tosin~P Adewumi, Foteini Liwicki, and Marcus Liwicki. 2020{\natexlab{b}}.
\newblock Exploring swedish \& english fasttext embeddings for ner with the
  transformer.
\newblock \emph{arXiv preprint arXiv:2007.16007}.

\bibitem[{Alkhaled et~al.(2023)Alkhaled, Adewumi, and
  Sabah~Sabry}]{alkhaled2023bipol}
Lama Alkhaled, Tosin Adewumi, and Sana Sabah~Sabry. 2023.
\newblock Bipol: A novel multi-axes bias evaluation metric with explainability
  for nlp.
\newblock \emph{Manuscript}.

\bibitem[{Antoniak and Mimno(2021)}]{antoniak2021bad}
Maria Antoniak and David Mimno. 2021.
\newblock Bad seeds: Evaluating lexical methods for bias measurement.
\newblock In \emph{Proceedings of the 59th Annual Meeting of the Association
  for Computational Linguistics and the 11th International Joint Conference on
  Natural Language Processing (Volume 1: Long Papers)}, pages 1889--1904.

\bibitem[{Bassignana et~al.(2018)Bassignana, Basile, and
  Patti}]{bassignana2018hurtlex}
Elisa Bassignana, Valerio Basile, and Viviana Patti. 2018.
\newblock Hurtlex: A multilingual lexicon of words to hurt.
\newblock In \emph{5th Italian Conference on Computational Linguistics, CLiC-it
  2018}, volume 2253, pages 1--6. CEUR-WS.

\bibitem[{Bender et~al.(2021)Bender, Gebru, McMillan-Major, and
  Shmitchell}]{bender2021dangers}
Emily~M Bender, Timnit Gebru, Angelina McMillan-Major, and Shmargaret
  Shmitchell. 2021.
\newblock On the dangers of stochastic parrots: Can language models be too big?
\newblock In \emph{Proceedings of the 2021 ACM Conference on Fairness,
  Accountability, and Transparency}, pages 610--623.

\bibitem[{Biewald(2020)}]{wandb}
Lukas Biewald. 2020.
\newblock \href {https://www.wandb.com/} {Experiment tracking with weights and
  biases}.
\newblock Software available from wandb.com.

\bibitem[{Chandrabose et~al.(2021)Chandrabose, Chakravarthi
  et~al.}]{chandrabose2021overview}
Aravindan Chandrabose, Bharathi~Raja Chakravarthi, et~al. 2021.
\newblock An overview of fairness in data--illuminating the bias in data
  pipeline.
\newblock In \emph{Proceedings of the First Workshop on Language Technology for
  Equality, Diversity and Inclusion}, pages 34--45.

\bibitem[{Clark et~al.(2019)Clark, Lee, Chang, Kwiatkowski, Collins, and
  Toutanova}]{clark-etal-2019-boolq}
Christopher Clark, Kenton Lee, Ming-Wei Chang, Tom Kwiatkowski, Michael
  Collins, and Kristina Toutanova. 2019.
\newblock \href {https://doi.org/10.18653/v1/N19-1300} {{B}ool{Q}: Exploring
  the surprising difficulty of natural yes/no questions}.
\newblock In \emph{Proceedings of the 2019 Conference of the North {A}merican
  Chapter of the Association for Computational Linguistics: Human Language
  Technologies, Volume 1 (Long and Short Papers)}, pages 2924--2936,
  Minneapolis, Minnesota. Association for Computational Linguistics.

\bibitem[{Clark et~al.(2020)Clark, Luong, Le, and Manning}]{clark2020electra}
Kevin Clark, Minh-Thang Luong, Quoc~V Le, and Christopher~D Manning. 2020.
\newblock Electra: Pre-training text encoders as discriminators rather than
  generators.
\newblock \emph{arXiv preprint arXiv:2003.10555}.

\bibitem[{Cryan et~al.(2020)Cryan, Tang, Zhang, Metzger, Zheng, and
  Zhao}]{10.1145/3313831.3376488}
Jenna Cryan, Shiliang Tang, Xinyi Zhang, Miriam Metzger, Haitao Zheng, and
  Ben~Y. Zhao. 2020.
\newblock \href {https://doi.org/10.1145/3313831.3376488} {Detecting gender
  stereotypes: Lexicon vs. supervised learning methods}.
\newblock In \emph{Proceedings of the 2020 CHI Conference on Human Factors in
  Computing Systems}, CHI '20, page 1–11, New York, NY, USA. Association for
  Computing Machinery.

\bibitem[{De~Marneffe et~al.(2019)De~Marneffe, Simons, and
  Tonhauser}]{de2019commitmentbank}
Marie-Catherine De~Marneffe, Mandy Simons, and Judith Tonhauser. 2019.
\newblock The commitmentbank: Investigating projection in naturally occurring
  discourse.
\newblock In \emph{proceedings of Sinn und Bedeutung}, volume~23, pages
  107--124.

\bibitem[{Dhamala et~al.(2021)Dhamala, Sun, Kumar, Krishna, Pruksachatkun,
  Chang, and Gupta}]{Dhamala2021}
Jwala Dhamala, Tony Sun, Varun Kumar, Satyapriya Krishna, Yada Pruksachatkun,
  Kai-Wei Chang, and Rahul Gupta. 2021.
\newblock \href
  {https://www.amazon.science/publications/bold-dataset-and-metrics-for-measuring-biases-in-open-ended-language-generation}
  {Bold: Dataset and metrics for measuring biases in open-ended language
  generation}.
\newblock In \emph{ACM FAccT 2021}.

\bibitem[{Dhar and Shamir(2021)}]{dhar2021evaluation}
Sanchari Dhar and Lior Shamir. 2021.
\newblock Evaluation of the benchmark datasets for testing the efficacy of deep
  convolutional neural networks.
\newblock \emph{Visual Informatics}, 5(3):92--101.

\bibitem[{Feng et~al.(2018)Feng, Fu, Dong, Guo, and Li}]{feng2018multistage}
Bo~Feng, Qiang Fu, Mianxiong Dong, Dong Guo, and Qiang Li. 2018.
\newblock Multistage and elastic spam detection in mobile social networks
  through deep learning.
\newblock \emph{IEEE Network}, 32(4):15--21.

\bibitem[{He et~al.(2021)He, Liu, Gao, and Chen}]{he2021deberta}
Pengcheng He, Xiaodong Liu, Jianfeng Gao, and Weizhu Chen. 2021.
\newblock \href {https://openreview.net/forum?id=XPZIaotutsD} {Deberta:
  Decoding-enhanced bert with disentangled attention}.
\newblock In \emph{International Conference on Learning Representations}.

\bibitem[{Heron(2009)}]{heron2009technologies}
Simon Heron. 2009.
\newblock Technologies for spam detection.
\newblock \emph{Network Security}, 2009(1):11--15.

\bibitem[{Levesque et~al.(2012)Levesque, Davis, and
  Morgenstern}]{levesque2012winograd}
Hector Levesque, Ernest Davis, and Leora Morgenstern. 2012.
\newblock The winograd schema challenge.
\newblock In \emph{Thirteenth international conference on the principles of
  knowledge representation and reasoning}.

\bibitem[{Liu et~al.(2019)Liu, Ott, Goyal, Du, Joshi, Chen, Levy, Lewis,
  Zettlemoyer, and Stoyanov}]{liu2019roberta}
Yinhan Liu, Myle Ott, Naman Goyal, Jingfei Du, Mandar Joshi, Danqi Chen, Omer
  Levy, Mike Lewis, Luke Zettlemoyer, and Veselin Stoyanov. 2019.
\newblock Roberta: A robustly optimized bert pretraining approach.
\newblock \emph{arXiv preprint arXiv:1907.11692}.

\bibitem[{Maddox(2004)}]{maddox2004perspectives}
Keith~B Maddox. 2004.
\newblock Perspectives on racial phenotypicality bias.
\newblock \emph{Personality and Social Psychology Review}, 8(4):383--401.

\bibitem[{Mehrabi et~al.(2021)Mehrabi, Morstatter, Saxena, Lerman, and
  Galstyan}]{mehrabi2021survey}
Ninareh Mehrabi, Fred Morstatter, Nripsuta Saxena, Kristina Lerman, and Aram
  Galstyan. 2021.
\newblock A survey on bias and fairness in machine learning.
\newblock \emph{ACM Computing Surveys (CSUR)}, 54(6):1--35.

\bibitem[{Monsen and J{\"o}nsson(2021)}]{monsen2021method}
Julius Monsen and Arne J{\"o}nsson. 2021.
\newblock A method for building non-english corpora for abstractive text
  summarization.
\newblock In \emph{Proceedings of CLARIN Annual Conference}.

\bibitem[{Nadeem et~al.(2021)Nadeem, Bethke, and
  Reddy}]{nadeem-etal-2021-stereoset}
Moin Nadeem, Anna Bethke, and Siva Reddy. 2021.
\newblock \href {https://doi.org/10.18653/v1/2021.acl-long.416} {{S}tereo{S}et:
  Measuring stereotypical bias in pretrained language models}.
\newblock In \emph{Proceedings of the 59th Annual Meeting of the Association
  for Computational Linguistics and the 11th International Joint Conference on
  Natural Language Processing (Volume 1: Long Papers)}, pages 5356--5371,
  Online. Association for Computational Linguistics.

\bibitem[{Nangia et~al.(2020)Nangia, Vania, Bhalerao, and
  Bowman}]{nangia-etal-2020-crows}
Nikita Nangia, Clara Vania, Rasika Bhalerao, and Samuel~R. Bowman. 2020.
\newblock \href {https://doi.org/10.18653/v1/2020.emnlp-main.154}
  {{C}row{S}-pairs: A challenge dataset for measuring social biases in masked
  language models}.
\newblock In \emph{Proceedings of the 2020 Conference on Empirical Methods in
  Natural Language Processing (EMNLP)}, pages 1953--1967, Online. Association
  for Computational Linguistics.

\bibitem[{Nosek et~al.(2002)Nosek, Banaji, and Greenwald}]{nosek2002harvesting}
Brian~A Nosek, Mahzarin~R Banaji, and Anthony~G Greenwald. 2002.
\newblock Harvesting implicit group attitudes and beliefs from a demonstration
  web site.
\newblock \emph{Group Dynamics: Theory, Research, and Practice}, 6(1):101.

\bibitem[{Nozza et~al.(2021)Nozza, Bianchi, and Hovy}]{nozza2021honest}
Debora Nozza, Federico Bianchi, and Dirk Hovy. 2021.
\newblock Honest: Measuring hurtful sentence completion in language models.
\newblock In \emph{The 2021 Conference of the North American Chapter of the
  Association for Computational Linguistics: Human Language Technologies}.
  Association for Computational Linguistics.

\bibitem[{Paullada et~al.(2021)Paullada, Raji, Bender, Denton, and
  Hanna}]{paullada2021data}
Amandalynne Paullada, Inioluwa~Deborah Raji, Emily~M Bender, Emily Denton, and
  Alex Hanna. 2021.
\newblock Data and its (dis) contents: A survey of dataset development and use
  in machine learning research.
\newblock \emph{Patterns}, 2(11):100336.

\bibitem[{Rudinger et~al.(2018{\natexlab{a}})Rudinger, Naradowsky, Leonard, and
  Van~Durme}]{rudinger-etal-2018-gender}
Rachel Rudinger, Jason Naradowsky, Brian Leonard, and Benjamin Van~Durme.
  2018{\natexlab{a}}.
\newblock \href {https://doi.org/10.18653/v1/N18-2002} {Gender bias in
  coreference resolution}.
\newblock In \emph{Proceedings of the 2018 Conference of the North {A}merican
  Chapter of the Association for Computational Linguistics: Human Language
  Technologies, Volume 2 (Short Papers)}, pages 8--14, New Orleans, Louisiana.
  Association for Computational Linguistics.

\bibitem[{Rudinger et~al.(2018{\natexlab{b}})Rudinger, Naradowsky, Leonard, and
  {Van Durme}}]{rudinger-EtAl:2018:N18}
Rachel Rudinger, Jason Naradowsky, Brian Leonard, and Benjamin {Van Durme}.
  2018{\natexlab{b}}.
\newblock Gender bias in coreference resolution.
\newblock In \emph{Proceedings of the 2018 Conference of the North American
  Chapter of the Association for Computational Linguistics: Human Language
  Technologies}, New Orleans, Louisiana. Association for Computational
  Linguistics.

\bibitem[{Sap et~al.(2020)Sap, Gabriel, Qin, Jurafsky, Smith, and
  Choi}]{sap-etal-2020-social}
Maarten Sap, Saadia Gabriel, Lianhui Qin, Dan Jurafsky, Noah~A. Smith, and
  Yejin Choi. 2020.
\newblock \href {https://doi.org/10.18653/v1/2020.acl-main.486} {Social bias
  frames: Reasoning about social and power implications of language}.
\newblock In \emph{Proceedings of the 58th Annual Meeting of the Association
  for Computational Linguistics}, pages 5477--5490, Online. Association for
  Computational Linguistics.

\bibitem[{Stanley(1977)}]{stanley1977paradigmatic}
Julia~Penelope Stanley. 1977.
\newblock Paradigmatic woman: The prostitute.
\newblock \emph{Papers in language variation}, pages 303--321.

\bibitem[{Subramanian et~al.(2021)Subramanian, Han, Baldwin, Cohn, and
  Frermann}]{subramanian2021evaluating}
Shivashankar Subramanian, Xudong Han, Timothy Baldwin, Trevor Cohn, and Lea
  Frermann. 2021.
\newblock Evaluating debiasing techniques for intersectional biases.
\newblock \emph{arXiv preprint arXiv:2109.10441}.

\bibitem[{Szumilas(2010)}]{szumilas2010explaining}
Magdalena Szumilas. 2010.
\newblock Explaining odds ratios.
\newblock \emph{Journal of the Canadian academy of child and adolescent
  psychiatry}, 19(3):227.

\bibitem[{Tan and Celis(2019)}]{tan2019assessing}
Yi~Chern Tan and L~Elisa Celis. 2019.
\newblock Assessing social and intersectional biases in contextualized word
  representations.
\newblock \emph{Advances in Neural Information Processing Systems}, 32.

\bibitem[{Tiedemann and Thottingal(2020)}]{TiedemannThottingal:EAMT2020}
J{\"o}rg Tiedemann and Santhosh Thottingal. 2020.
\newblock {OPUS-MT} — {B}uilding open translation services for the {W}orld.
\newblock In \emph{Proceedings of the 22nd Annual Conferenec of the European
  Association for Machine Translation (EAMT)}, Lisbon, Portugal.

\bibitem[{Wang et~al.(2019)Wang, Pruksachatkun, Nangia, Singh, Michael, Hill,
  Levy, and Bowman}]{NEURIPS2019_4496bf24}
Alex Wang, Yada Pruksachatkun, Nikita Nangia, Amanpreet Singh, Julian Michael,
  Felix Hill, Omer Levy, and Samuel Bowman. 2019.
\newblock \href
  {https://proceedings.neurips.cc/paper/2019/file4496bf24afe7fab6f046bf4923da8de6-Paper.pdf}
  {Superglue: A stickier benchmark for general-purpose language understanding
  systems}.
\newblock In \emph{Advances in Neural Information Processing Systems},
  volume~32. Curran Associates, Inc.

\bibitem[{Wolf et~al.(2020)Wolf, Debut, Sanh, Chaumond, Delangue, Moi, Cistac,
  Rault, Louf, Funtowicz, Davison, Shleifer, von Platen, Ma, Jernite, Plu, Xu,
  Le~Scao, Gugger, Drame, Lhoest, and Rush}]{wolf-etal-2020-transformers}
Thomas Wolf, Lysandre Debut, Victor Sanh, Julien Chaumond, Clement Delangue,
  Anthony Moi, Pierric Cistac, Tim Rault, Remi Louf, Morgan Funtowicz, Joe
  Davison, Sam Shleifer, Patrick von Platen, Clara Ma, Yacine Jernite, Julien
  Plu, Canwen Xu, Teven Le~Scao, Sylvain Gugger, Mariama Drame, Quentin Lhoest,
  and Alexander Rush. 2020.
\newblock \href {https://doi.org/10.18653/v1/2020.emnlp-demos.6} {Transformers:
  State-of-the-art natural language processing}.
\newblock In \emph{Proceedings of the 2020 Conference on Empirical Methods in
  Natural Language Processing: System Demonstrations}, pages 38--45, Online.
  Association for Computational Linguistics.

\bibitem[{Xue et~al.(2021)Xue, Constant, Roberts, Kale, Al-Rfou, Siddhant,
  Barua, and Raffel}]{xue-etal-2021-mt5}
Linting Xue, Noah Constant, Adam Roberts, Mihir Kale, Rami Al-Rfou, Aditya
  Siddhant, Aditya Barua, and Colin Raffel. 2021.
\newblock \href {https://doi.org/10.18653/v1/2021.naacl-main.41} {m{T}5: A
  massively multilingual pre-trained text-to-text transformer}.
\newblock In \emph{Proceedings of the 2021 Conference of the North American
  Chapter of the Association for Computational Linguistics: Human Language
  Technologies}, pages 483--498, Online. Association for Computational
  Linguistics.

\bibitem[{Yudkowsky et~al.(2008)}]{yudkowsky2008artificial}
Eliezer Yudkowsky et~al. 2008.
\newblock Artificial intelligence as a positive and negative factor in global
  risk.
\newblock \emph{Global catastrophic risks}, 1(303):184.

\bibitem[{Zhao et~al.(2018)Zhao, Wang, Yatskar, Ordonez, and
  Chang}]{zhao-etal-2018-gender}
Jieyu Zhao, Tianlu Wang, Mark Yatskar, Vicente Ordonez, and Kai-Wei Chang.
  2018.
\newblock \href {https://doi.org/10.18653/v1/N18-2003} {Gender bias in
  coreference resolution: Evaluation and debiasing methods}.
\newblock In \emph{Proceedings of the 2018 Conference of the North {A}merican
  Chapter of the Association for Computational Linguistics: Human Language
  Technologies, Volume 2 (Short Papers)}, pages 15--20, New Orleans, Louisiana.
  Association for Computational Linguistics.

\end{thebibliography}


\appendix


\subsection{Experiment}
\label{experiment}


\subsubsection*{Dictionary of lists for \acrshort{roberta} on \acrshort{boolq}}

\{'gender': [{' she ': 23, ' her ': 17, ' woman ': 2, ' lady ': 1, ' female ': 6, ' girl ': 1, ' skirt ': 0, ' madam ': 0, ' gentlewoman ': 0, ' madame ': 0, ' dame ': 0, ' gal ': 0, ' maiden ': 0, ' maid ': 0, ' damsel ': 0, ' senora ': 0, ' lass ': 0, ' beauty ': 0, ' ingenue ': 0, ' belle ': 0, ' doll ': 0, ' señora ': 0, ' senorita ': 0, ' lassie ': 0, ' ingénue ': 0, ' miss ': 0, ' mademoiselle ': 0, ' señorita ': 0, ' babe ': 0, ' girlfriend ': 0, ' lover ': 0, ' mistress ': 0, ' ladylove ': 0, ' inamorata ': 0, ' gill ': 0, ' old ': 2, ' beloved ': 0, ' dear ': 0, ' sweetheart ': 0, ' sweet ': 0, ' flame ': 2, ' love ': 5, ' valentine ': 0, ' favorite ': 1, ' moll ': 0, ' darling ': 0, ' honey ': 0, ' significant ': 0, ' wife ': 3, ' wifey ': 0, ' missus ': 0, ' helpmate ': 0, ' helpmeet ': 0, ' spouse ': 0, ' bride ': 1, ' partner ': 0, ' missis ': 0, ' widow ': 0, ' housewife ': 0, ' mrs ': 0, ' matron ': 0, ' soul ': 3, ' mate ': 1, ' housekeeper ': 0, ' dowager ': 0, ' companion ': 0, ' homemaker ': 0, ' consort ': 0, ' better half ': 0, ' hausfrau ': 0, ' stay-at-home ': 0}, {' he ': 80, ' him ': 49, ' boy ': 3, ' man ': 1, ' male ': 10, ' guy ': 1, ' masculine ': 0, ' virile ': 0, ' manly ': 0, ' man-sized ': 0, ' hypermasculine ': 0, ' macho ': 0, ' mannish ': 0, ' manlike ': 0, ' man-size ': 0, ' hairy-chested ': 0, ' butch ': 0, ' ultramasculine ': 0, ' boyish ': 0, ' tomboyish ': 0, ' hoydenish ': 0, ' amazonian ': 0, ' gentleman ': 0, ' dude ': 0, ' fellow ': 0, ' cat ': 2, ' gent ': 0, ' fella ': 0, ' lad ': 0, ' bloke ': 0, ' bastard ': 0, ' joe ': 0, ' chap ': 0, ' chappie ': 0, ' hombre ': 0, ' galoot ': 0, ' buck ': 0, ' joker ': 3, ' mister ': 0, ' jack ': 8, ' sir ': 0, ' master ': 1, ' buddy ': 0, ' buster ': 0}], 'racial':...
\}


\end{document}